\documentclass[sigconf]{acmart}

\usepackage{amsmath,amsfonts}
\usepackage{algorithmic}
\usepackage{graphicx}
\usepackage{cite}

\usepackage{subcaption}
\usepackage{threeparttable}
\usepackage{tabularx}
\usepackage{booktabs}

\usepackage{textcomp}
\usepackage{xcolor}

\def \ie {\emph{i.e.}}

\newcommand{\ourmodel}{ALADIN}    
\AtBeginDocument{%
  \providecommand\BibTeX{{%
    \normalfont B\kern-0.5em{\scshape i\kern-0.25em b}\kern-0.8em\TeX}}}

\setcopyright{acmcopyright}
\copyrightyear{2022}
\acmYear{2022}
\acmDOI{XXXXXXX.XXXXXXX}

\acmConference[CBMI 2022]{International Conference on Content-based Multimedia Indexing}{Sept. 14-16, 2022}{Graz, Austria}
\acmPrice{15.00}
\acmISBN{978-1-4503-XXXX-X/18/06}

\begin{document}

\title{ALADIN: Distilling Fine-grained Alignment Scores for Efficient Image-Text Matching and Retrieval}


\author{Nicola Messina}
\affiliation{%
  \institution{ISTI-CNR}
  \city{Pisa}
  \country{Italy}}
\email{nicola.messina@isti.cnr.it}

\author{Matteo Stefanini}
\affiliation{%
  \institution{Univ. of Modena and Reggio Emilia}
  \city{Modena}
  \country{Italy}}
\email{matteo.stefanini@unimore.it}

\author{Marcella Cornia}
\affiliation{%
  \institution{Univ. of Modena and Reggio Emilia}
  \city{Modena}
  \country{Italy}}
\email{marcella.cornia@unimore.it}

\author{Lorenzo Baraldi}
\affiliation{%
  \institution{Univ. of Modena and Reggio Emilia}
  \city{Modena}
  \country{Italy}}
\email{lorenzo.baraldi@unimore.it}

\author{Fabrizio Falchi}
\affiliation{%
  \institution{ISTI-CNR}
  \city{Pisa}
  \country{Italy}}
\email{fabrizio.falchi@isti.cnr.it}

\author{Giuseppe Amato}
\affiliation{%
  \institution{ISTI-CNR}
  \city{Pisa}
  \country{Italy}}
\email{giuseppe.amato@isti.cnr.it}

\author{Rita Cucchiara}
\affiliation{%
  \institution{Univ. of Modena and Reggio Emilia}
  \city{Modena}
  \country{Italy}}
\email{rita.cucchiara@unimore.it}

\renewcommand{\shortauthors}{N. Messina, et al.}

\begin{abstract}
Image-text matching is gaining a leading role among tasks involving the joint understanding of vision and language. In literature, this task is often used as a pre-training objective to forge architectures able to jointly deal with images and texts. Nonetheless, it has a direct downstream application: \textit{cross-modal retrieval}, which consists in finding images related to a given query text or vice-versa.
Solving this task is of critical importance in cross-modal search engines.
Many recent methods proposed effective solutions to the image-text matching problem, mostly using recent large vision-language (VL) Transformer networks. However, these models are often computationally expensive, especially at inference time. This prevents their adoption in large-scale cross-modal retrieval scenarios, where results should be provided to the user almost instantaneously. In this paper, we propose to fill in the gap between effectiveness and efficiency by proposing an ALign And DIstill Network (\ourmodel). \ourmodel\ first produces high-effective scores by aligning at fine-grained level images and texts. Then, it learns a shared embedding space -- where an efficient kNN search can be performed -- by distilling the relevance scores obtained from the fine-grained alignments. We obtained remarkable results on MS-COCO, showing that our method can compete with state-of-the-art VL Transformers while being almost 90 times faster. The code for reproducing our results is available at \url{https://github.com/mesnico/ALADIN}.
\end{abstract}

\ccsdesc[500]{Information systems~Information retrieval}
\ccsdesc[500]{Information systems~Multimedia and multimodal retrieval}
\ccsdesc[500]{Computing methodologies~Neural networks}
\ccsdesc[500]{Computing methodologies~Computer vision}
\ccsdesc[500]{Computing methodologies~Visual content-based indexing and retrieval}
\ccsdesc[500]{Computing methodologies~Matching}

\keywords{cross-modal retrieval, image-text matching, vision-and-language.}


\maketitle

\section{Introduction}
\label{sec:intro}
With the growing strength of deep learning methods and the availability of large-scale data, multi-modal processing has become one of the most promising research topics. In particular, most of the focus is placed on the joint processing of images and natural language sentences. By understanding the hidden semantic connections between a text and an image, many works in literature solved challenging multi-modal problems, such as image captioning~\citep{Anderson2018bottomup,cornia2020meshed,stefanini2022show} or visual question answering~\citep{Anderson2018bottomup,zhou2020unified,banerjee2021weakly}.
Among these tasks, \textit{image-text matching} has crucial importance~\citep{kiros2014unifying,vsepp2018faghri,cornia2020unified,messina2021fine,messina2021towards}: it consists of outputting a relevance score for each given (image, text) pair, where the score is high if the image is relevant to the text and low otherwise. Although this task is usually employed as a vision-language pre-training objective, it is crucial for cross-modal \textit{image-text retrieval}, which usually consists of two sub-tasks: \textit{image retrieval}, where we want images relevant to a given text, and \textit{text retrieval}, where we ask for sentences better describing an input image. Efficiently and effectively solving these retrieval tasks is strategically important in modern cross-modal search engines.

Many state-of-the-art models for image-text matching, like Oscar~\citep{li2020oscar} or UNITER~\citep{uniter2020}, comprise large and deep multi-modal vision-language (VL) Transformers with early fusion, which are computationally expensive, especially during the inference phase. In fact, during inference, all the (image, text) pairs from the test set should be forwarded through the multi-modal Transformer to obtain the relevance scores. This is clearly unfeasible in large datasets and unusable in large-scale retrieval scenarios, where the system latency should be as small as possible.

For achieving such a performance objective, many approaches in the literature project image and text embeddings in a common space where similarity is measured through simple dot products. This allows the introduction of an \textit{offline} phase, in which all the dataset items are encoded and stored, and an \textit{online} phase in which only the query is forwarded through the network and compared with all the offline-stored elements. Although these approaches are very efficient, they are usually not sufficiently effective as the ones relying on early modality fusion using large VL Transformers.

In the light of these observations, in this paper we propose an ALign And DIstill Network model (\textit{\ourmodel}), which exploits the knowledge acquired by large VL Transformers to craft an efficient yet effective model for image-text retrieval. In particular, we employ late fusion approaches so that the two visual and textual pipelines are kept separated until the final matching phase. The first objective consists of \textit{aligning} image regions with sentence words, using a simple yet effective alignment head. Then, a common visual-textual embedding space is learned by distilling the scores from the alignment head using a learning-to-rank objective. In this case, we use the learned alignment scores as ground-truth (teacher) scores. 

We show that, on the widely used MS-COCO dataset, the alignment scores can reach results comparable with large joint vision-language models such as UNITER and OSCAR, while being far more efficient, especially during inference. On the other hand, the distilled scores used to learn the common space can defeat previous common space methods on the same dataset, opening the way toward metric-based indexing for large-scale retrieval.

To sum up, in this paper, we propose the following contributions:
\begin{itemize}
    \item We employ two instances of a pre-trained VL Transformer as a backbone for extracting separate visual and textual features. 
    \item We adopt a simple yet effective alignment method for producing high-quality scores instead of the poorly-scalable output of large joint VL Transformers.
    \item We create an informative embedding space by framing the problem as a learning-to-rank task and distilling the final scores using the scores in output from the alignment head.
\end{itemize}
\section{Related Work}
\label{sec:related}

In the last years, many works tackled the image-text matching task. The work in~\citep{vsepp2018faghri} paved the way for the common space approach for cross-modal matching. They showed the effectiveness of the hinge-based triplet ranking loss with hard-negative mining. Many works followed their footsteps~\citep{messina2021fine, messina2021transformer,li2019,stefanini2021novel,qu2020context,wen2020learning}, trying out BERT~\citep{devlin2019bert} as a text extractor other than a simple GRU and showing the effectiveness of region-based features~\citep{Anderson2018bottomup} as visual representation. After the success of BERT-like models in Natural Language Processing~\citep{devlin2019bert,lewis2019bart,liu2019roberta}, many works tried to employ the Transformer Encoder to jointly process images and text, like VilBERT~\citep{lu2019vilbert}, OSCAR~\citep{li2020oscar}, VL-BERT~\citep{Su2020VL-BERT}, or VinVL~\citep{zhang2021vinvl}. These methods tackle image-text matching as a binary classification problem, where an (image, sentence) pair is input to the complex Transformer architecture which is trained to predict the probability that the sentence relates to the image. Although these architectures are very effective, they are computationally expensive at inference time, as they need to process every (image, sentence) pair to obtain the scores on the whole test set. For this reason, many methods keep the visual and textual pipelines separated, without cross-talking between them~\citep{messina2021fine,messina2021transformer,huang2018image,sarafianos2019adversarial,wen2020learning}. Doing so, they can be forwarded independently at inference time, at the cost of losing effectiveness. Our work is inspired by the recent success of knowledge distillation~\citep{anil2018large,barraco2022camel,caron2021emerging,xie2020self,zhou2020more}, used to transfer knowledge from a large model to a smaller and more efficient one. We propose to use scores distillation to learn a visual-textual common space, employing the knowledge acquired by a pre-trained VL Transformer. In this case, the knowledge distillation is framed as a learning-to-rank problem~\citep{cao2007learning,pobrotyn2020context,bruch2021alternative}, widely used in literature but, as far as we know, never used for distilling cross-modal scores.

\section{Proposed Method}
\label{sec:method}
The proposed architecture is composed of two different stages.
The first stage, which we refer to as \textit{backbone}, is composed of the layers of a pre-trained large vision-language transformer -- VinVL~\citep{zhang2021vinvl}, an extension to the powerful OSCAR model~\citep{li2020oscar}. In the backbone, the language and the visual paths do not interact through cross-attention mechanisms so that the features from the two modalities can be extracted independently at inference time.

\begin{figure*}[t]
  \centering
  \includegraphics[width=.98\linewidth]{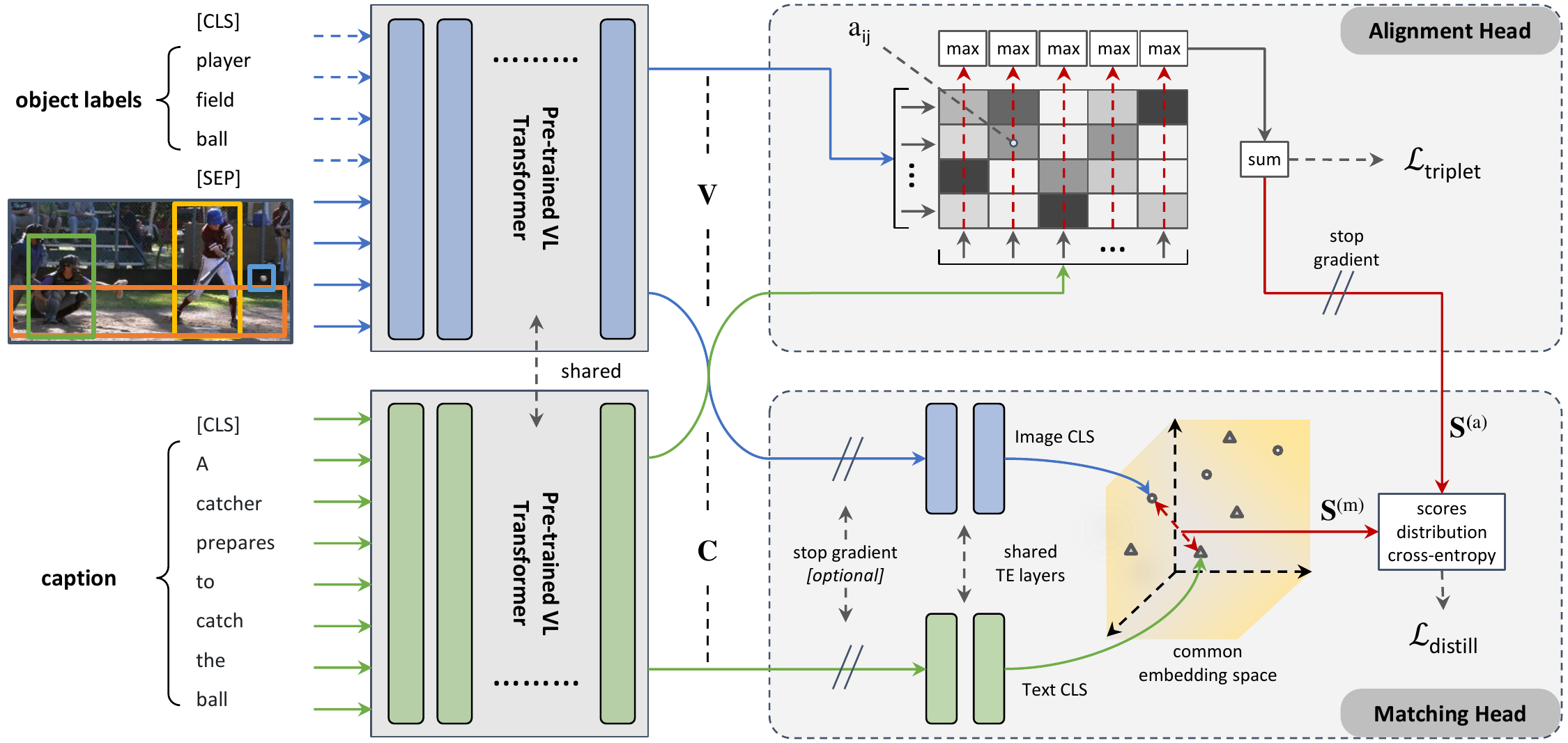}
  \caption{Overview of our architecture. The backbone extracts visual and textual features that are used in both the matching and alignment heads. The matching head is trained by distilling the scores using the ones coming from the alignment head.}
  \label{fig:architecture}
\end{figure*}

The second stage, instead, is composed of two separate \textit{heads}: the \textit{alignment} head and \textit{matching} head. The alignment head is used to pre-train the network to efficiently align the visual and the textual concepts in a fine-grained manner, as done in TERAN~\citep{messina2021fine}. Differently, the matching head is used to construct an informative cross-modal common space, that can be used to efficiently represent images and text as fixed-length vectors for use in large-scale retrieval. The scores from the matching head are distilled using the scores from the alignment head as guidance. The overall architecture is shown in Figure~\ref{fig:architecture}.

In the following, we dive into the building blocks of the architecture -- \ie, the backbone, the alignment head, and the matching head.

\subsection{Vision-Language Backbone}
As the backbone for feature extraction, we use the pre-trained layers from VinVL~\citep{zhang2021vinvl}, an extension to the large-scale vision-language OSCAR model~\citep{li2020oscar}.
Our goal is to obtain suitable vectorial representations for the image $\mathcal{V}$ and the text $\mathcal{C}$ in input. In particular, we employ the model pre-trained on the image-text retrieval task. The authors used a binary classification head on top of the CLS token of the output sequence, and the model is trained to predict if the input images and textual sentences are related or not.

In our use case, the visual and textual pipelines should be separated, so that they can be forwarded independently at inference time.
For this reason, we use two instances of the VinVL architecture, in a shared-weights configuration to forward the two modalities independently, as shown in Figure~\ref{fig:architecture}. 

As in~\citep{zhang2021vinvl}, we use as visual tokens both the visual features extracted from object regions\footnote{\url{https://github.com/microsoft/scene\_graph\_benchmark}} and their labels, and the two sub-sequences are separated by a SEP token.
In the end, the outputs from the last layers of the disentangled VinVL architecture are two sequences, $\boldsymbol{V} = \{\boldsymbol{v}_\text{cls}, \boldsymbol{v}_1, \boldsymbol{v}_2,\ldots,\boldsymbol{v}_N\}$, representing the image $\mathcal{V}$, and $\boldsymbol{C} = \{\boldsymbol{c}_\text{cls}, \boldsymbol{c}_1, \boldsymbol{c}_2,\ldots,\boldsymbol{c}_M\}$, representing the text $\mathcal{C}$.
Note that, in both sequences, the first element is the CLS token, used to collect representative information for the whole image or text.

\subsection{Alignment Head}
The alignment head comprises a similarity matrix that computes the fine-grained relevances between the visual tokens $\boldsymbol{V}$ and textual tokens $\boldsymbol{C}$. The fine-grained similarities are then pooled to obtain the final global relevance between the image and the text. In particular, we use a formulation similar to the one used in TERAN~\citep{messina2021fine}.
Specifically, the features in output from the backbone are used to compute a visual-textual tokens alignment matrix $\boldsymbol{A} \in \mathbb{R}^{n \times m}$, built as follows: 
\begin{equation}
    \boldsymbol{A} = a_{ij}^{kl} = \text{cosine}(\boldsymbol{v}_i, \boldsymbol{c}_j) = \frac{\boldsymbol{v}_i^T \boldsymbol{c}_j}{\| \boldsymbol{v}_i \| \| \boldsymbol{c}_j \|} \qquad i \in g_k, j \in g_l, \label{eq:alignment-matrix}
\end{equation}
where $g_k$ is the set of indexes of the region features from the $k$-th image and $g_l$ is the set of indexes of the words from the $l$-th sentence. 
At this point, the similarities $s_{kl}$ between the image $k$ and the caption $l$ are computed by pooling the similarity matrix $\boldsymbol{A}$ along dimensions $(i, j)$ through an appropriate pooling function. Guided by~\citep{messina2021fine}, we use the \textit{max-over-regions sum-over-words} policy, which computes the following final similarity score:

\begin{equation}
    \boldsymbol{S}^\text{(a)} = s^\text{(a)}_{kl} = \sum_{j \in g_l}\max_{i \in g_k} {A_{ij}}.
\end{equation}

The dot-product similarity used to compute $\boldsymbol{A}$ in Eq. \ref{eq:alignment-matrix} resembles the computation of the cross-attention between visual and textual tokens. The difference boils down to the interaction between the visual and textual pipelines, which happens only at the very end of the whole architecture. This \textit{late cross-attention} makes the sequences $\boldsymbol{V}$ and $\boldsymbol{C}$ cacheable, eliminating the need to forward the whole architecture whenever a new query -- either visual or textual -- is issued to the system. The computation of $\boldsymbol{S^\text{(a)}}$, involving only simple non-parametric operations, is very efficient and can be easily implemented on GPU to obtain high inference speeds.

The loss function used to force this network to produce suitable similarities $s$ for each (image, text) pair is the \textit{hinge-based triplet ranking loss}, used in previous works~\citep{vsepp2018faghri,li2019,messina2021fine}. Formally,
\begin{align}
    \mathcal{L}_\text{triplet} = \sum_{k,l} \max_{{l}'} [\alpha + s_{kl'} - s_{kl}]_+ + \max_{{k}'} [\alpha + s_{k'l} - s_{kl}]_+,  \label{eq:triplet-loss}
\end{align}
where $s_{kl}$ is the similarity estimated between image $k$ and caption $l$, and $[x]_+ \equiv \max(0, x)$; the values ${k}',{l}'$ are the indexes of the image and caption hard negatives found in the mini-batch as done in~\citep{vsepp2018faghri}, and $\alpha$ is a margin that defines the minimum separation that should hold between positive and negative pairs.

Given that the alignment head is directly connected to the backbone, we fine-tuned the backbone on this new alignment objective. More details on the training procedure are reported in Section~\ref{sec:training}.

\subsection{Matching Head}
\label{sec:matching-head}
The matching head uses the same sequences $\boldsymbol{V}$ and $\boldsymbol{C}$ given from the backbone and employs them to produce the features $\boldsymbol{\tilde{v}} \in \mathbb{R}^{d}$ for the image $\mathcal{V}$ and $\boldsymbol{\tilde{c}} \in \mathbb{R}^{d}$ for the caption $\mathcal{C}$. These representations are forced to lay in the same $d$-dimensional embedding space. In this space, $k$-neirest-neighbor search can be efficiently computed --- using metric space approaches or inverted files --- to quickly retrieve images given a textual query or vice-versa.
Specifically, we forward $\boldsymbol{V}$ and $\boldsymbol{C}$ through a 2-layer Transformer Encoder (TE):
\begin{align}
\boldsymbol{\bar{V}} = \text{TE}(\boldsymbol{V}); \qquad
\boldsymbol{\bar{C}} = \text{TE}(\boldsymbol{C}).
\end{align}
As in~\citep{messina2021transformer}, the TE shares the weights among the two modalities, and the final vectors encoding the whole image and caption are the CLS tokens in output from the TE layers: $\boldsymbol{\tilde{v}} = \boldsymbol{\bar{V}}[0] = \boldsymbol{\bar{v}}_\text{cls}$ and $\boldsymbol{\tilde{c}} = \boldsymbol{\bar{C}}[0] = \boldsymbol{\bar{c}}_\text{cls}$.
The final relevances are simply computed as the cosine similarities between the the vector $\boldsymbol{\tilde{v}}_k$ from the $k$-th image and $\boldsymbol{\tilde{s}}_l$ from the $l$-th sentence: $\boldsymbol{S}^\text{(m)} = s^\text{(m)}_{kl} = \text{cosine}(\boldsymbol{\tilde{v}}_k,\boldsymbol{\tilde{s}}_l)$.

In principle, we could optimize the common space using the same hinge-based triplet ranking loss in Eq. \ref{eq:triplet-loss} already used to train the alignment head. 
Instead, in the light of the good effectiveness-efficiency trade-off of the alignment head, we propose to learn a distribution for $\boldsymbol{S}^\text{(m)}$ using the previously-learned $\boldsymbol{S}^\text{(a)}$ as teachers.

Specifically, we frame the problem of distilling the distribution of $\boldsymbol{S}^\text{(m)}$ from $\boldsymbol{S}^\text{(a)}$ as a \textit{learning-to-rank} problem.
We employ the mathematical framework developed in the ListNet approach~\citep{cao2007learning}, which models the probability of an object being ranked at the top, given the scores of all the objects. Differently from this framework, here we need to optimize for two different entangled distributions: the distribution of text-image similarities when sentences are used as queries, and the distribution of image-text similarities when instead images are used as queries.
In particular, given a textual query $k$ and an image query $l$, the probabilities of the image $i$ and text $j$ to be the top-one elements respectively with respect to $\boldsymbol{S}^\text{(a)}$ are:
\begin{align}
P_{\boldsymbol{S}^\text{(a)}}(i) = \frac{\exp(s_{ik}^\text{(a)})}{\sum_{t=1}^{B} \exp(s_{tk}^\text{(a)})};  P_{\boldsymbol{S}^\text{(a)}}(j) = \frac{\exp(s_{lj}^\text{(a)})}{\sum_{t=1}^{B} \exp(s_{tj}^\text{(a)})}
\end{align}
where $B$ is the batch size, as the learning procedure is confined to the images and sentences in the current batch. Therefore, during training, only $B$ images are retrieved using the query $k$, and $B$ textual elements are retrieved using the query $l$.
Similarly, an analogous probability can be defined over $\boldsymbol{S}^\text{(m)}$:
\begin{align}
P_{\boldsymbol{S}^\text{(m)}}(i) = \frac{\exp(\tau s_{ik}^\text{(m)})}{\sum_{t=1}^{B} \exp(\tau s_{tk}^\text{(m)})};  P_{\boldsymbol{S}^\text{(m)}}(j) = \frac{\exp(\tau s_{lj}^\text{(m)})}{\sum_{t=1}^{B} \exp(\tau s_{tj}^\text{(m)})}
\end{align}
where $\tau$ is a temperature hyper-parameter which compensates for the fact that $\boldsymbol{S}^\text{(m)}$ ranges in [0, 1]. We empirically found that $\tau=6.0$ works well in practice.
The final matching loss can be formulated as the cross-entropy between the $P_{\boldsymbol{S}^\text{(a)}}$ and $P_{\boldsymbol{S}^\text{(m)}}$ probabilities, for both the image-to-text and text-to-image cases.
\begin{align}
    \mathcal{L}_{\text{distill}} = - \sum_{i=1}^B P_{\boldsymbol{s}^\text{(a)}(i)} \log(P_{\boldsymbol{s}^\text{(m)}}(i)) - \sum_{j=1}^B P_{\boldsymbol{s}^\text{(a)}(j)} \log(P_{\boldsymbol{s}^\text{(m)}}(j))
\end{align}

Notice that accurate and dense teacher scores are needed to obtain a good estimate of the teacher distributions $P_{\boldsymbol{s}^\text{(a)}}(i)$ and $P_{\boldsymbol{s}^\text{(a)}}(j)$. This partly motivates our choice of first researching an effective and efficient alignment head that could output the scores to be used as ground-truth for the matching head.

\subsection{Training}
\label{sec:training}

During the training phase, we initially respect the following constraints: (a) the backbone is finetuned only when training the alignment head, and (b) the gradients do not flow backward through $\boldsymbol{S}^\text{(a)}$ when training the matching head (as depicted in Figure \ref{fig:architecture} through the \textit{stop-gradient} indication). The constraint (b) comes from the fact that the scores $\boldsymbol{S}^\text{(a)}$ are used as teacher scores. Therefore, they should not modify the weights of the backbone, because it is assumed that the backbone is already trained with the alignment head. Given these constraints, we train the network in two steps. First, we train the alignment head by updating the backbone weights using $\mathcal{L}_\text{triplet}$ (\textbf{\ourmodel\ A/ft.} in the experiments). Then, we freeze the backbone and we learn the matching head by updating the weights of the 2-layer Transformer Encoder using $\mathcal{L}_\text{distill}$ (\textbf{\ourmodel\ D} in the experiments). Note that the formalism \textit{X/ft.} signifies that the gradients coming from that head loss X are used to finetune the backbone. Possible head losses are X=\{T, D, A\} for T=triplet, D=distillation, and A=alignment, where T and D come from the matching head, while A from the alignment head. When \textit{/ft.} is omitted, it means that the backbone remains frozen.

We explore also the joint training of the two heads. Specifically, we relax constraint (a), so that gradients coming from the two heads can update the backbone. Sticking to the previous formalism, we refer to this experiment as \textbf{\ourmodel\ A/ft. + D/ft.}. Nevertheless, when directly applying this training schema, we experienced some instabilities. If the alignment head --- working as a teacher for the matching head --- is not warmed-up, it can not initially provide good teacher scores. The consequence is that noisy gradients backpropagate through the matching head and interfere with the finetuning of the backbone.
For this reason, we warmup the backbone by pre-training it with the alignment loss $\mathcal{L}_\text{triplet}$ (as in the \textit{\ourmodel\ A/ft.} setup). 

\section{Experiments}
\label{sec:experiments}

In this section, we report detailed results for validating our approach.
In addition to the training setups described in \ref{sec:training}, we consider two more schemes as baselines: \textbf{\ourmodel\ T} trains the matching head using the standard hinge-based triplet ranking loss without distillation, starting from a pre-trained backbone (\ie~\ourmodel\ A/ft.) and leaving it fixed; similarly, \textbf{\ourmodel\ T/ft.} lacks the alignment head and the backbone is finetuned only with the gradients from the matching head.

\subsection{Dataset and Metrics}
We perform our experiments on the widely-used MS-COCO dataset, which contains a large corpus of images scraped from the web. Each image is annotated with 5 textual descriptions. We follow the splits introduced by~\citep{karpathy2015alignment}, which reserves 113,287 images for training, 5,000 for validating, and 5,000 for testing. In literature, a smaller test set comprising only 1,000 images is often used. For a fair comparison, we report the results on both 5K and 1K test sets. In the case of 1K images, the results are computed by performing a 5-fold cross-validation and averaging the results.

As commonly done to evaluate cross-modal retrieval models~\citep{vsepp2018faghri,li2019,qi2020imagebert,lu2019vilbert,lee2019}, we use the recall@$k$ metric for evaluating the ability of our model to correctly retrieve relevant texts or images. Specifically, the recall@$k$ measures the percentage of queries able to retrieve the correct item among the first $k$ results.


\subsection{Alignment Head Results}

\newcolumntype{C}{>{\centering\arraybackslash}X}
\newcolumntype{R}{D{,}{\pm}{1.6}}
\newcolumntype{L}{>{\raggedright\arraybackslash}p{3cm}}
\renewcommand{\arraystretch}{0.9}
\begin{table*}[htbp]
\small
\caption{Experiment results using scores from the alignment head. The comparison is performed with entangled visual-textual Transformer models.}
\begin{center}
\begin{tabular}{Lccccccccccccc}
\toprule
& & \multicolumn{6}{c}{1K Test Set} & \multicolumn{6}{c}{5K Test Set} \\
\cmidrule(lr){3-8} \cmidrule(lr){9-14}
& & \multicolumn{3}{c}{Text Retrieval} & \multicolumn{3}{c}{Image Retrieval} & \multicolumn{3}{c}{Text Retrieval} & \multicolumn{3}{c}{Image Retrieval} \\
\cmidrule(lr){3-5} \cmidrule(lr){6-8} \cmidrule(lr){9-11} \cmidrule(lr){12-14}
\textbf{Model} & Training Data & \multicolumn{1}{c}{$k=1$} & \multicolumn{1}{c}{$k=5$} & \multicolumn{1}{c}{$k=10$}
& \multicolumn{1}{c}{$k=1$} & \multicolumn{1}{c}{$k=5$} & \multicolumn{1}{c}{$k=10$} & \multicolumn{1}{c}{$k=1$} & \multicolumn{1}{c}{$k=5$} & \multicolumn{1}{c}{$k=10$} & \multicolumn{1}{c}{$k=1$} & \multicolumn{1}{c}{$k=5$} & \multicolumn{1}{c}{$k=10$} \\
\midrule
12-in-1~\citep{lu202012} & 4.4M & - & - & - & 65.2 & 91.0 & 96.2 & - & - & - & - & - & - \\
VilBERT~\citep{lu2019vilbert} & 3.1M & - & - & - & 58.2 & 84.9 & 91.5 & - & - & - & - & - & - \\
Unicoder-VL~\citep{li2020unicoder} & 3.8M & 84.3 & 97.3 & 99.3& 69.7 & 93.5 & 97.2 & 62.3 & 87.1 & 92.8 & 46.7 & 76.0 & 85.3 \\
UNITER (Base)~\citep{uniter2020} & 5.6M & - & - & - & - & - & - & 63.3 & 87.0 & 93.1 & 48.4 & 76.7 & 85.9 \\
OSCAR (Base)~\citep{li2020oscar} & 6.5M & - & - & - & - & - & - & 70.0 & 91.1 & 95.5 & 54.0 & 80.8 & 88.5 \\
VinVL (Base)~\citep{zhang2021vinvl} & 8.9M & - & - & - & - & - & - & \textbf{74.6} & \textbf{92.6} & \textbf{96.3} & \textbf{58.1} & \textbf{83.2} & \textbf{90.1} \\
\midrule
\textbf{\ourmodel\ A/ft.} & 8.9M & \textbf{88.1} & \textbf{99.1} & \textbf{99.7} & \textbf{75.4} & \textbf{95.2} & 97.9 & 70.0 & 90.7 & 95.6 & 54.4 & 81.0 & 88.6 \\
\textbf{\ourmodel\ A/ft. + D/ft.} & 8.9M & 87.6 & 98.5 & \textbf{99.7} & 75.0 & \textbf{95.2} & \textbf{98.0} & 69.9 & 91.3 & 95.7 & 54.7 & 81.0 & 88.7 \\
\bottomrule
\end{tabular}
\label{tab:alignment-results}
\end{center}
\end{table*}

\newcolumntype{C}{>{\centering\arraybackslash}X}
\newcolumntype{R}{D{,}{\pm}{1.6}}
\newcolumntype{L}{>{\raggedright\arraybackslash}p{3cm}}
\begin{table*}[htbp]
\small
\caption{Experimental results using scores from the matching head. The comparison is performed with methods using disentangled visual-textual pipelines.}
\begin{center}
\begin{tabular}{Lccccccccccccc}
\toprule
& & \multicolumn{6}{c}{1K Test Set} & \multicolumn{6}{c}{5K Test Set} \\
\cmidrule(lr){3-8} \cmidrule(lr){9-14}
& & \multicolumn{3}{c}{Text Retrieval} & \multicolumn{3}{c}{Image Retrieval} & \multicolumn{3}{c}{Text Retrieval} & \multicolumn{3}{c}{Image Retrieval} \\
\cmidrule(lr){3-5} \cmidrule(lr){6-8} \cmidrule(lr){9-11} \cmidrule(lr){12-14}
\textbf{Model} & Training Data & \multicolumn{1}{c}{$k=1$} & \multicolumn{1}{c}{$k=5$} & \multicolumn{1}{c}{$k=10$}
& \multicolumn{1}{c}{$k=1$} & \multicolumn{1}{c}{$k=5$} & \multicolumn{1}{c}{$k=10$} & \multicolumn{1}{c}{$k=1$} & \multicolumn{1}{c}{$k=5$} & \multicolumn{1}{c}{$k=10$} & \multicolumn{1}{c}{$k=1$} & \multicolumn{1}{c}{$k=5$} & \multicolumn{1}{c}{$k=10$} \\
\midrule
TERN~\citep{messina2021transformer} & 0.6M & 65.5 & 91.0 & 96.5 & 54.5 & 86.9 & 94.2 & 40.2 & 71.1 & 81.9 & 31.4 & 62.5 & 75.3 \\
SAEM (ens.)~\citep{wu2019learning} & 0.6M & 71.2 & 94.1 & 97.7 & 57.8 & 88.6 & 94.9 & - & - & - & - & - & - \\
CAMERA (ens.)~\citep{qu2020context} & 0.6M & 77.5 & 96.3 & 98.8 & 63.4 & 90.9 & 95.8 & 55.1 & 82.9 & 91.2 & 40.5 & 71.7 & 82.5\\
TERAN (ens.)~\citep{messina2021fine} & 0.6M & 80.2 & 96.6 & 99.0 & 67.0 & 92.2 & 96.9 & 59.3 & 85.8  & 92.4 & 45.1 & 74.6 & 84.4 \\
DSRAN (w. BERT)~\citep{wen2020learning} & 0.6M & 80.6 & 96.7 & 98.7 & 64.5 & 90.8 & 95.8 & 57.9 & 85.3 & 92.0 & 41.7 & 72.7 & 82.8 \\
\midrule
\textbf{\ourmodel\ T} & 8.9M & 79.2 & 96.7 & 99.1 & 68.9 & 92.8 & 96.6 & 57.9 & 84.8 & 91.8 & 46.0 & 74.8 & 84.1 \\
\textbf{\ourmodel\ D} & 8.9M & 83.1 & 97.4 & 99.3 & 70.5 & 93.6 & 97.3 & 62.7 & 87.5 & 93.5 & 47.4 & 76.2 & 85.4 \\
\textbf{\ourmodel\ T/ft.} & 8.9M & \textbf{84.9} & \textbf{98.5} & 99.6 & 71.9 & 93.8 & 97.0 & 63.6 & 87.4 & 93.5 & 49.7 & 77.7 & 86.3 \\
\textbf{\ourmodel\ A/ft. + D/ft.} & 8.9M & 84.7 & 98.0 & \textbf{99.8} & \textbf{72.7} & \textbf{94.5} & \textbf{97.5} & \textbf{64.9} & \textbf{88.6} & \textbf{94.5} & \textbf{51.3} & \textbf{79.2} & \textbf{87.5} \\
\midrule
\midrule
CLIP (0-shot)~\citep{radford2021learning} & \textit{0.4B} & - & - & - & - & - & - & \textit{58.4} & \textit{81.5} & \textit{88.1} & \textit{37.8} & \textit{62.4} & \textit{72.2} \\
ALIGN~\citep{jia2021scaling} & \textit{1.8B} & - & - & - & - & - & - & \textit{77.0} & \textit{93.5} & \textit{96.9} & \textit{59.9} & \textit{83.3} & \textit{89.8} \\
\bottomrule
\end{tabular}
\label{tab:matching-results}
\end{center}
\end{table*}

We first compare the results obtained with our alignment head against some recent methods comprising large-scale pre-trained Transformer models (Table~\ref{tab:alignment-results}). We consider only the \textit{Base} versions and not the \textit{Large} ones, for hardware limitations. For a fair comparison, we initialize our backbone with the weights of VinVL Base~\citep{zhang2021vinvl}. 
Notice that, at test time, all the reported models except ours need to compute a number of network forward steps in the order of $O(n^2 r)$, where $n$ is the number of images and $r$ is the number of sentences associated to each image ($r=5$ in case of MS-COCO). In fact, due to cross-attention links between visual and textual pipelines, intermediate representations cannot be cached for being reused with a different query. Instead, given the disentangled pipelines, our model enables caching of the image and text features in output from the backbone for speeding up the retrieval with never seen queries, with a number of network forward steps in the order of $O(n + nr)$. As we can notice from Table \ref{tab:alignment-results}, this disentanglement comes at the cost of a slight reduction of the overall effectiveness, as we can notice by comparing our approach to the VinVL model. Nevertheless, our model \ourmodel\ A/ft. can perfectly compete, and partially overtake, all the previous entangled visual-textual Transformer models on both image and sentence retrieval tasks. From the results on the \ourmodel\ A/ft. + D/ft. model, we can notice that when the distillation loss is also active the alignment scores are pretty comparable to \ourmodel\ A/ft. In particular, on the 5K test set, we observe slight improvements in both image and sentence retrieval. This evidence suggests that the distillation loss has the collateral effect of regularizing its own teacher scores, as done in recent works on self-distillation~\citep{zhang2019your,caron2021emerging}.

\subsection{Matching Head Results}
We compare the common space created from our matching head with other disentangled methods using similar approaches. The results are shown in Table \ref{tab:matching-results}. As explained above, for comparison we report also the matching head directly trained using the hinge-based triplet loss (\ourmodel\ T and \ourmodel\ T/ft.) without distilling the scores from the alignment head. Furthermore, for completeness, we report also the results from the recent methods CLIP (0-shot)~\citep{radford2021learning} and ALIGN~\citep{jia2021scaling}. Although the comparison with CLIP (0-shot) may result unfair, we decided to stick with the results obtained by the authors of the original paper, to avoid all the intricacies deriving from the hyper-parameter tuning phase needed for a satisfactory fine-tuning stage. However, these models use from $100\times$ to $1000\times$ more training data, so we exclude them from the analysis.

All of our methods outperform the previous models, notably surpassing TERAN~\citep{messina2021fine}, the method that introduced the alignment matrix used in the alignment head. Concerning the experiments that non-finetune the backbones (\ourmodel\ T and \ourmodel\ D), we argue that scores distillation helps, especially in the recall@1, where we observe an improvement of about 8\% and 2\% on sentence and image retrieval respectively for the 5K test set. 
We obtain the best results by using our model \ourmodel\ A/ft. + D/ft., which jointly trains the alignment and distillation heads by also finetuning the backbone with the respective gradients. The alignment scores from this setup already proved to be effective in Table~\ref{tab:alignment-results}. The distilled scores in output from the matching head follow the same trend, obtaining the best results on the 5K test set.

\begin{figure}[t]
  \centering
  \includegraphics[width=\linewidth]{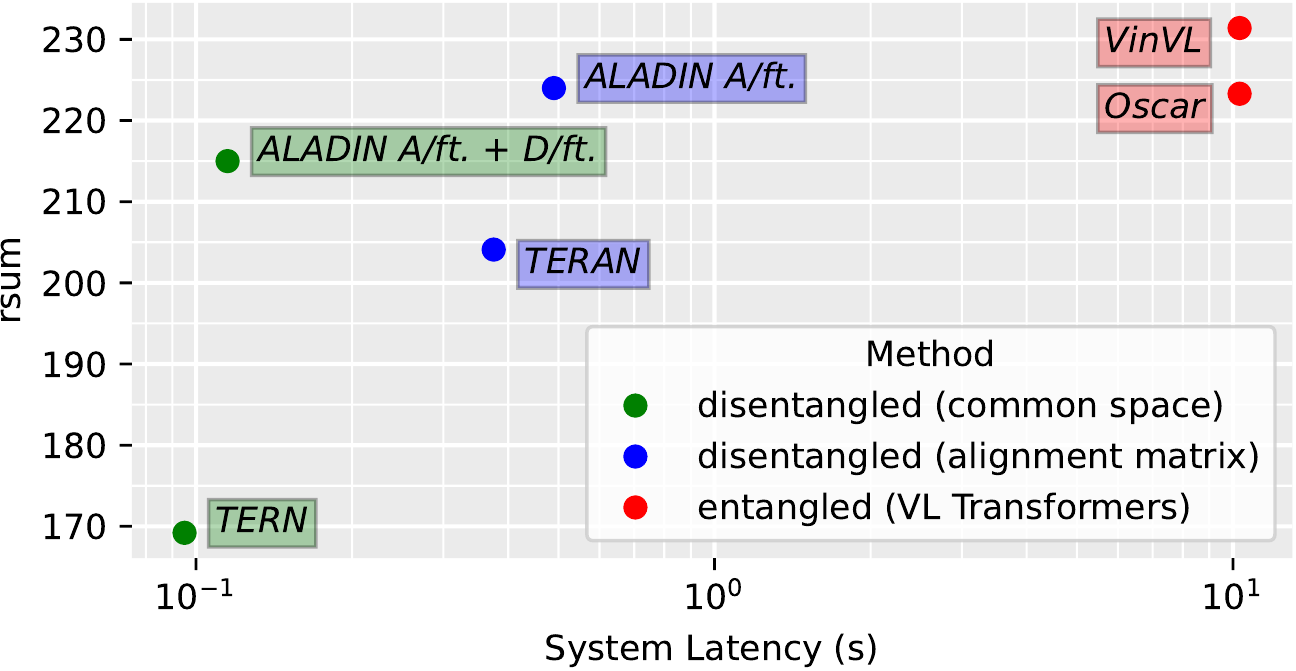}
  \caption{Effectiveness vs efficiency. We report effectiveness as the sum of recall values on the image retrieval (rsum), and efficiency as the time needed to search the 5K test images.}
  \label{fig:effectiveness-vs-efficiency}
\end{figure}

\subsection{Effectiveness vs Efficiency}
To better show the advantage of our model in terms of computing times, in Figure~\ref{fig:effectiveness-vs-efficiency} we plot the effectiveness vs the efficiency of our approach compared with other methods. We address image-retrieval on the 1K test set, and we report the sum of the recall values (rsum) versus the average time needed to solve a textual query. These experiments are run on a system equipped with an RTX 2080Ti and an AMD Ryzen 7 1700 Eight-Core Processor. As we can notice, the scores from the alignment head (\textit{\ourmodel\ A/ft.}) can directly compete with VL Transformer models, although being almost 20 times faster. Notably, the scores computed on the distilled space from \textit{\ourmodel\ A/ft. + D/ft.} obtain a speedup of almost 90$\times$, with a rsum loss of only about 7\% with respect to VinVL.
Therefore, the proposed models help fill the gap between efficiency and effectiveness -- \ie, the top left zone of the diagram.

Considering the efficiency-effectiveness trade-offs of both the alignment and matching heads, the whole architecture could be deployed in real application scenarios in a two-stage configuration: first, the faster matching head proposes relevant candidates using k-NN search on the common space; then, the candidates are re-ranked using the scores from the alignment head. This pipeline would enable the alignment head, which is slower but more effective, to contribute to the final ranking while keeping the whole system highly scalable.

\section{Conclusions}
In this paper, we presented an efficient and effective architecture for visual-textual cross-modal retrieval. Specifically, we proposed to learn an alignment score by independently forwarding the visual and the textual pipelines using a state-of-the-art VL Transformer as a backbone. Then, we used the scores produced by the alignment head to learn a visual-textual common space, which can produce easily indexable fixed-length features. Specifically, we approached the problem using a learn-to-rank distillation objective, which empirically demonstrated its effectiveness over the standard hinge-based triplet ranking loss to optimize the common space. The experiments conducted on MS-COCO confirmed the validity of our approach. The results demonstrated that this method helps fill the gap between effectiveness and efficiency, enabling this system to be deployed in large-scale cross-modal retrieval scenarios. 

\section*{Acknowledgments}
This work has been partially supported by AI4CHSites CNR4C program (CUP B15J19001040004), by AI4Media under GA 951911, by the ``Artificial Intelligence for Cultural Heritage (AI4CH)'' project, co-funded by the Italian Ministry of Foreign Affairs and International Cooperation, and by the PRIN project ``CREATIVE: CRoss-modal understanding and gEnerATIon of Visual and tExtual content'' (CUP B87G22000460001), co-funded by the Italian Ministry of University and Research.

\bibliographystyle{ACM-Reference-Format}
\bibliography{biblio}


\end{document}